\documentclass[conference]{IEEEtran}
\IEEEoverridecommandlockouts
\usepackage{cite}
\usepackage{amsmath,amssymb,amsfonts}
\usepackage{algorithmic}
\usepackage{graphicx}
\usepackage{textcomp}
\usepackage{xcolor}
\def\BibTeX{{\rm B\kern-.05em{\sc i\kern-.025em b}\kern-.08em
    T\kern-.1667em\lower.7ex\hbox{E}\kern-.125emX}}

\begin{document}

\title{Segmentation over Complexity: Evaluating Ensemble and Hybrid Approaches for Anomaly Detection in Industrial Time Series}

\author{
\IEEEauthorblockN{Emilio Mastriani, Alessandro Costa, Federico Incardona, Kevin Munari, Sebastiano Spinello}
\IEEEauthorblockA{
\textit{INAF, Osservatorio Astrofisico di Catania, Catania, Italy} \\
\{emilio.mastriani, alessandro.costa, federico.incardona, kevin.munari, sebastiano.spinello\}@inaf.it
}
}
\maketitle

\begin{abstract}
In this study, we investigate the effectiveness of advanced feature engineering and hybrid model architectures for anomaly detection in a multivariate industrial time series, focusing on a steam turbine system. We evaluate the impact of change point-derived statistical features, clustering-based substructure representations, and hybrid learning strategies on detection performance. Despite their theoretical appeal, these complex approaches consistently underperformed compared to a simple Random Forest + XGBoost ensemble trained on segmented data. The ensemble achieved an AUC-ROC of 0.976, F1-score of 0.41, and 100\% early detection within the defined time window. Our findings highlight that, in scenarios with highly imbalanced and temporally uncertain data, model simplicity combined with optimized segmentation can outperform more sophisticated architectures, offering greater robustness, interpretability, and operational utility.
\end{abstract}


\section{Introduction}
In recent years, anomaly detection in time series has become a critical challenge in industrial applications [1]. The timely identification of anomalous behaviors can prevent critical failures, reduce downtime, and significantly improve overall operational efficiency. However, accurate anomaly detection is complicated by the multivariate nature of sensor data and the inherent uncertainty in temporal labels provided by domain experts. Often, precise information about exact failure dates is unavailable, with only indicative time intervals alternating between normal and anomalous states available. Furthermore, declared failure periods typically represent a small percentage of the available data, making anomaly identification particularly challenging [2].
To address these challenges, various segmentation techniques have been proposed to reduce temporal uncertainty and improve detection model effectiveness [3].Among these, Change Point Detection (CPD) methods [4], such as ChangeFinder[5], have proven especially effective in identifying significant transitions between different operational states, providing valuable preprocessing for supervised machine learning models. The adoption of segmentation approaches can enhance prediction accuracy compared to traditional anomaly detection methods. ChangeFinder, as an online unsupervised algorithm for anomaly and change point detection, demonstrates notable efficiency in identifying sudden changes in statistical properties, including shifts in mean or variance.
In predictive maintenance applications, recent studies have emphasized that time series segmentation can provide crucial information for detecting transitions between normal and anomalous states [6]. Concurrently, heterogeneous ensemble learning, which combines models with complementary characteristics, has shown promising results in enhancing anomaly detection robustness and accuracy. However, model effectiveness significantly depends on feature quality and the ability to correctly isolate relevant state changes [5].
\subsection{The Allure and Peril of Complexity in ML Research}
In machine learning research, there exists an inherent bias toward complex solutions—the assumption that more features, advanced algorithms, and sophisticated architectures will inevitably yield superior results. This pursuit of complexity often leads researchers down paths of increasingly intricate feature engineering and model architectures, sometimes without rigorous validation against established baselines [7]
\subsection{Our Established Baseline and Research Challenge}
In our previous work [8] we demonstrated that simple data segmentation combined with a Random Forest and XGBoost ensemble achieved state-of-the-art performance for our anomaly detection problem, attaining an AUC-ROC of 0.9760 and F1-score of 0.41. This straightforward approach effectively addressed the dataset's inherent challenges, including temporal uncertainty and class imbalance. This success naturally raises a critical research question: Could advanced feature engineering and hybrid model architectures push performance beyond this established baseline? Specifically, would techniques such as change point statistics, advanced clustering algorithms, and sophisticated hybrid models deliver measurable improvements over our proven simple ensemble approach?
\subsection{Study Context and Dataset}
The equipment studied in this work is a steam turbine connected to an electric generator within a fully digitalized industrial plant. The steam turbine converts pressure drop from high-pressure steam (HP) to medium-pressure steam (MP) into electrical energy, effectively recovering energy. During turbine unavailability, steam can be diverted through a dedicated valve that reduces steam pressure from HP to MP. The turbine's electricity production is directly linked to plant utility steam demand and, consequently, to the refining center's production level [9].
The dataset comprises 70 variables (features) containing 1,124,820 data points. Training data covers the period from July 9, 2022, to August 3, 2023, while test data spans from September 1, 2023, to November 22, 2024. The confirmed anomaly range extends from August 11, 2024, to August 17, 2024—a 7-day interval representing approximately 1.56\% of the total 448-day test dataset duration.
The data frame object in this study is accompanied by a Normal Operating Condition (NoC) file identifying periods during which industry experts assessed the turbine as working under normal conditions. This file served two primary purposes: (a) identifying machine operating and idle periods as temporal sequence sets to quantify segmentation effectiveness, and (b) using the compressor's operating state (normal/anomalous) as the target feature during hybrid model training.
The dataset's natural imbalance, with predominant "normal" data compared to anomalous events, necessitated [8] prioritizing models capable of handling imbalanced situations before assessing segmentation technique contributions. This paper presents our comprehensive investigation into whether sophisticated approaches could surpass our established simple baseline, documenting both the methodological journey and its unexpected conclusions.

\section{Methods}

\subsection{Phase 1: Change Point Statistics Feature Engineering}
The first phase of our methodology aimed to enhance the segmented dataset by deriving statistical features from change point detection. The central objective was to capture the dynamics preceding structural transitions, which could potentially improve anomaly detection capabilities by providing richer temporal context.
To characterize pre-transition behavior, we introduced five features. \textbf{mean\_score\_pre\_cp} measures the average anomaly score prior to the most recent change point, indicating the level of system instability before structural changes. \textbf{dist\_last\_cp} captures the temporal distance from the last change point, with lower values reflecting recent transitions and higher values denoting prolonged stability. \textbf{max\_score\_pre\_cp} records the maximum anomaly score before the last change point, highlighting peak deviations potentially signaling imminent faults. \textbf{std\_score\_pre\_cp} represents the standard deviation of pre-change point scores, reflecting local variability and instability. Finally, \textbf{cp\_freq} quantifies the frequency of change points within defined temporal windows, summarizing the system’s long-term stability patterns.

The enriched dataset was evaluated across multiple models, and the results are summarized in Table 1. These initial tests indicated a counterintuitive outcome: while some features appeared theoretically informative, the inclusion of all five features often led to a marked decrease in predictive performance. 
\begin{table}[htbp]
\centering
\caption{Model performance comparison across feature sets}
\footnotesize
\setlength{\tabcolsep}{3pt} 
\begin{tabular}{p{2.3cm} p{2.0cm} c c c}
\hline
\textbf{Model} & \textbf{Metric} & \textbf{Baseline} & \textbf{5 Features} & \textbf{3 Features} \\
\hline
Random Forest & AUC-ROC & 0.96 & 0.39 & 0.76 \\
              & Avg Precision & 0.16 & 0.01 & 0.04 \\
Isolation Forest & AUC-ROC & 0.6885 & 0.5131 & 0.5384 \\
                 & ETP (\%) & 91.96 & 51.19 & 53.87 \\
XGBoost & AUC-ROC & 0.8759 & 0.5955 & 0.6820 \\
        & ETP (\%) & 0.00 & 100.00 & 51.34 \\
One-Class SVM & Best AUC-ROC & 0.7823 & 0.9016 & 0.8833 \\
              & Best F1-score & 0.0308 & 0.1469 & 0.0308 \\
\hline
\end{tabular}
\end{table}

The observed decline in model performance when incorporating all five change point features prompted a detailed examination of their statistical distributions. As illustrated in Figure 1, this analysis revealed marked differences in discriminative potential among the features.
Both dist\_last\_cp and cp\_freq exhibited nearly identical distributions across classes, with long tails and minimal separation, suggesting low discriminative power. In contrast, mean\_score\_pre\_cp demonstrated clear separation between classes, with negative values (up to -50) predominating in one class and positive values clustered in the other (approximately 5–10), indicating a strong signal for pre-transition behavior.
Similarly, std\_score\_pre\_cp showed noticeable distinctions, with the False class clustered at low values and the True class exhibiting greater dispersion and extended tails, reflecting higher variability preceding change points. Finally, max\_score\_pre\_cp revealed a distinct distribution pattern, with the True class extending across long tails in both negative and positive directions, while the False class remained confined within a narrow range.
Based on what was observed, we decided to keep only the mean\_score\_pre\_cp, std\_score\_pre\_cp, and max\_score\_pre\_cp features. In order to validate the theoretical coherence of these features, we computed both the inter- and intra-segment variance of the three features as a direct quantitative measure of segment separability. The F-ratio measure, defined as \(F-ratio=\frac{Var(inter-segment)}{Var(intra-segment)}\) reflects the trade-off between the dispersion of segment centroids and the compactness of individual segments: higher values of F-ratio indicate well-separated and homogeneous segments. Table 2 summarizes the F-ratios. 
Exceptionally high F-ratios, ranging from 300,000 to 700,000, confirmed that the features maintained minimal variation within segments while exhibiting substantial differences between segments—precisely the desired characteristic for capturing state transitions.

\begin{table}[htbp]
\caption{Top Ranked Features Based on F-ratio}
\centering
\setlength{\tabcolsep}{3pt} 
\begin{tabular}{c l r}
\hline
\textbf{Rank} & \textbf{Feature} & \textbf{F-ratio} \\
\hline
1 & ElectricalEfficiency\_std\_score\_pre\_cp & 718,604.84 \\
2 & ElectricalEfficiency\_mean\_score\_pre\_cp & 639,643.32 \\
3 & V470-A165-A.pv\_mean\_score\_pre\_cp & 329,466.05 \\
4 & V470PT001.pv\_max\_score\_pre\_cp & 322,963.29 \\
\hline
\end{tabular}
\end{table}

That is why, only the three most discriminative features—mean\_score\_pre\_cp, std\_score\_pre\_cp, and max\_score\_pre\_cp—were retained for final testing. Despite their theoretical promise, this refined feature set did not surpass the baseline performance of the segmented dataset. For example, looking at Table 1, the Random Forest model improved from an AUC-ROC of 0.39 (all features) to 0.76 (top three features), yet remained well below the baseline value of 0.96.
In conclusion, although change point-derived statistical features were coherent and theoretically meaningful, their integration introduced additional noise without enhancing discriminative power.
\begin{figure}
    \centering
    \includegraphics[width=1\linewidth]{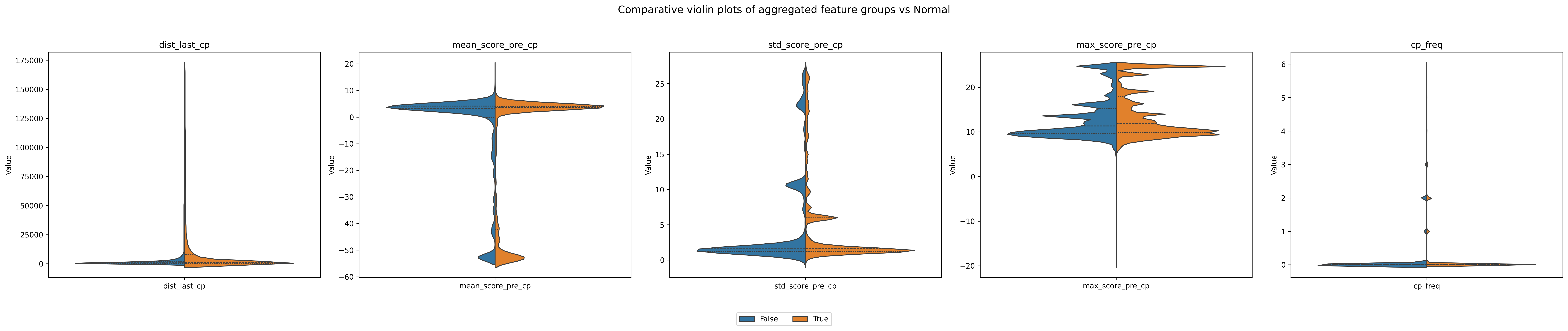}
    \caption{Violin plots of feature groups (dist\_last\_cp, mean\_score\_pre\_cp, std\_score\_pre\_cp, max\_score\_pre\_cp, cp\_freq) comparing Normal and anomalous samples, showing distribution differences and class-separating patterns.}
    \label{fig:placeholder}
\end{figure}

\subsection{Phase 2: Advanced Clustering}
In order to capture latent structural patterns within the time series and enhance the predictive potential of the dataset, an unsupervised clustering analysis was performed on each segment of the monitored variables. This approach aims to identify recurrent operational states and micro-clusters of homogeneous behavior that may act as precursors of anomalous or degraded conditions [10]. Each identified sub-cluster was added to the dataset as a categorical feature,  allowing subsequent models to exploit latent structural information not captured by the original variables. To ensure coverage of the main clustering paradigms, several representative algorithms were evaluated: KMeans (partition-based), Gaussian Mixture Models (probabilistic), BIRCH (hierarchical), OPTICS and HDBSCAN (density-based), and Mean Shift (mode-seeking). Together, these methods provide a balanced assessment across centroid, probabilistic, hierarchical, and density-driven strategies.

Clustering quality was assessed using three internal metrics---Silhouette Coefficient, Calinski--Harabasz (CH) Index, and Davies--Bouldin (DB) Index---computed per segment and averaged across all segments. The Silhouette measures intra-cluster cohesion and inter-cluster separation; the CH index favors configurations with high between-cluster variance and low within-cluster dispersion; the DB index penalizes overlapping clusters, with lower values indicating better structure.

Figure 2 shows that among the tested algorithms, KMeans, BIRCH, and Mean Shift achieved moderate Silhouette values \((\sim{0.55}) \) and acceptable DB scores, though their CH indices were artificially inflated, suggesting metric sensitivity to scale. GMM performed worst (Silhouette = 0.52; DB = 0.70), producing weakly separated clusters. Density-based methods performed best: OPTICS (Silhouette = 0.66; DB = 0.45) yielded robust, well-separated groups, while HDBSCAN achieved the highest overall quality (Silhouette = 0.69; DB = 0.44) with realistic CH scores. Overall, HDBSCAN proved the most effective, with OPTICS offering a strong alternative for datasets exhibiting variable local densities.

\begin{figure}
    \centering
    \includegraphics[width=0.75\linewidth]{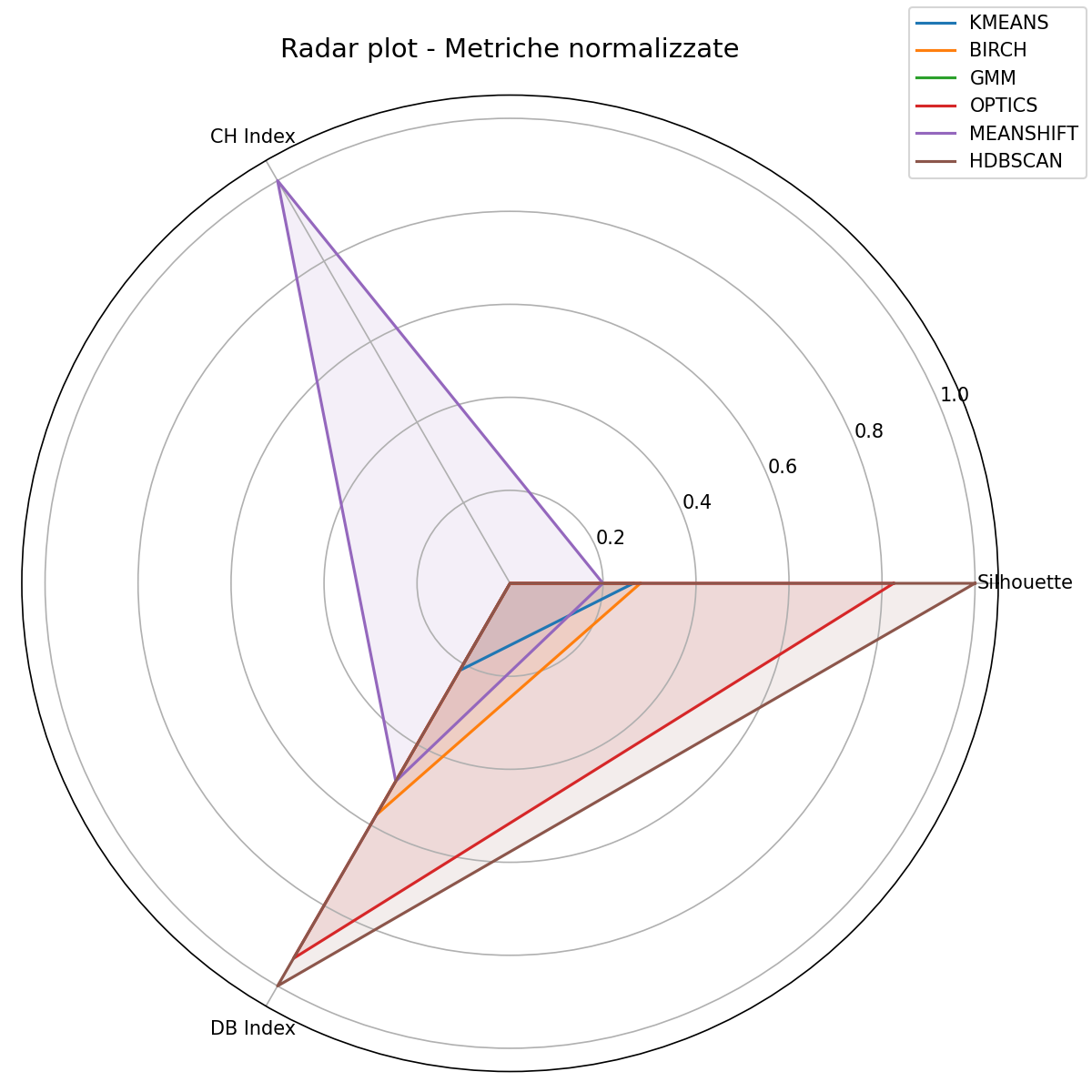}
    \caption{Radar plot of normalized clustering metrics (Silhouette, Calinski–Harabasz, and Davies–Bouldin) for all evaluated algorithms (KMeans, BIRCH, GMM, OPTICS, MeanShift, and HDBSCAN). The plot provides a visual comparison of each algorithm’s overall performance, with larger enclosed areas indicating superior clustering quality across the combined criteria.}
    \label{fig:placeholder}
\end{figure}
As mentioned before, the F-ratio measure reflects the trade-off between the dispersion of cluster centroids and the compactness of individual clusters: higher values of F-ratio indicate well-separated and homogeneous clusters.
Building upon this concept, the \(\Delta F\) index was introduced as a comparative measure between two density-based algorithms (OPTICS and HDBSCAN) defined as: \[\Delta F=F_{optics}-F_{hdbscan}\]
A positive \(\Delta F\) indicates that OPTICS achieves greater cluster separability, while a negative \(\Delta F \) suggests that HDBSCAN produces more compact and cohesive clusters.
This additional index enables a direct, quantitative comparison between the two methods in terms of their ability to balance inter-cluster distinctiveness and intra-cluster homogeneity.
Both F and \(\Delta F\) were computed for each cluster and subsequently the \(\Delta F\) has been integrated into the main dataset as new features. This integration allows the clustering structure to be explicitly represented in the data used for downstream modeling tasks, such as anomaly detection or degradation forecasting. 
In summary, the introduction of clustering-based features, combined with the evaluation of internal validation metrics and the \(\Delta F\) index, provides a systematic and data-driven approach for enriching the original dataset. This methodology enhances the interpretability of the latent structures within the data and improves the overall modeling robustness, particularly in complex, non-linear, and noisy industrial time-series environments [11].
\subsubsection{Comparative Analysis of Density-Based Segmentation: OPTICS vs. HDBSCAN}
A comparative analysis of F-ratio distributions obtained via OPTICS and HDBSCAN highlights complementary segmentation behaviors. OPTICS captures finer and more heterogeneous substructures, showing greater sensitivity to local density variations, whereas HDBSCAN forms fewer, more coherent clusters with discriminative power concentrated in a limited set of features.

Methodologically, OPTICS is well suited for exploratory analyses aimed at detecting micro-patterns or early anomaly precursors, while HDBSCAN provides stable, noise-filtered segmentations for confirmatory modeling. Combined, these approaches create a balanced framework in which OPTICS supports pattern discovery and HDBSCAN ensures robustness. Integrating such density-based segmentation with Random Forest– and permutation-based feature assessments effectively reveals both stable and transient dynamics in complex operational datasets.

Contrary to expectations, results remained lower than the baseline, as shown in Table 3.

\begin{table*}[t]
\centering
\caption{Performance comparison between segmented and segmented enriched datasets}
\resizebox{\textwidth}{!}{%
\begin{tabular}{lcc}
\hline
\textbf{Model} & \textbf{Segmented} & \textbf{Segmented Enriched + DF} \\
\hline
\textbf{Random Forest} & 
ROC-AUC: 0.96 \newline 
AP: 0.16 \newline 
TTD (mean): 0.00 \newline 
ETP: 672/672 (100\%) &
ROC-AUC: 0.54 \newline 
AP: 0.02 \newline 
TTD (mean): 0.00 \newline 
ETP: 672/672 (100\%) \\
\hline
\textbf{Isolation Forest} & 
AUC-ROC: 0.6885 \newline 
TTD (mean steps): 1.18 \newline 
ETP: 309/336 (91.96\%) &
AUC-ROC: 0.5322 \newline 
TTD (mean steps): 6.16 \newline 
ETP: 192/336 (57.14\%) \\
\hline
\textbf{XGBoost} & 
AUC-ROC: 0.8759 \newline 
Threshold: 0.9973 \newline 
TTD (mean): nan \newline 
ETP: 0/672 (0.00\%) &
AUC-ROC: 0.6810 \newline 
Threshold: 0.9868 \newline 
TTD (mean): 19.64 \newline 
ETP: 345/672 (51.34\%) \\
\hline
\textbf{K-Means} & 
AUC-ROC: 0.2177 \newline 
Threshold: -26372.31 \newline 
TTD (mean): 0.00 \newline 
ETP: 672/672 (100\%) &
AUC-ROC: 0.2177 \newline 
Threshold: -35382479.16 \newline 
TTD (mean): 0.00 \newline 
ETP: 672/672 (100\%) \\
\hline
\textbf{PCA} & 
AUC-ROC: 0.7823 \newline 
Threshold: 5.88$\times10^6$ \newline 
TTD (mean): 0.00 \newline 
ETP: 672/672 (100\%) &
AUC-ROC: 0.7823 \newline 
Threshold: 7.82$\times10^{12}$ \newline 
TTD (mean): 0.00 \newline 
ETP: 672/672 (100\%) \\
\hline
\textbf{One-Class SVM} & 
AUC-ROC: 0.7823 \newline 
F1-score: 0.0308 \newline 
TTD (mean): 0.00 \newline 
ETP: 672/672 (100\%) &
AUC-ROC: 0.8833 \newline 
F1-score: 0.0308 \newline 
TTD (mean): 0.00 \newline 
ETP: 672/672 (100\%) \\
\hline
\end{tabular}%
}
\label{tab:segmentation_comparison}
\end{table*}

\subsection{Phase 3: Feature Relevance Analysis: Random Forest and Segment-Level Permutation Importance}
Suspecting the introduction of noise due to the added features, we moved to identify the most informative predictors within operational segments by a dual-stage analytical strategy was employed. Initially, feature importance derived from a Random Forest model provided a global ranking of variables according to their overall discriminative power. Subsequently, to refine this analysis and mitigate potential inter-segment structural bias, Permutation Importance was computed within each segment. This segment-level evaluation quantified how local perturbations of individual features impacted the model’s predictive stability, ensuring that importance scores reflected genuine intra-cluster discriminative ability rather than merely global correlations.
Results summarized in Figure 3 and and Table 4 on-based analysis converge on a coherent set of features that dominate the model’s predictive capacity, although from complementary perspectives.

\begin{figure}
    \centering
    \includegraphics[width=1\linewidth]{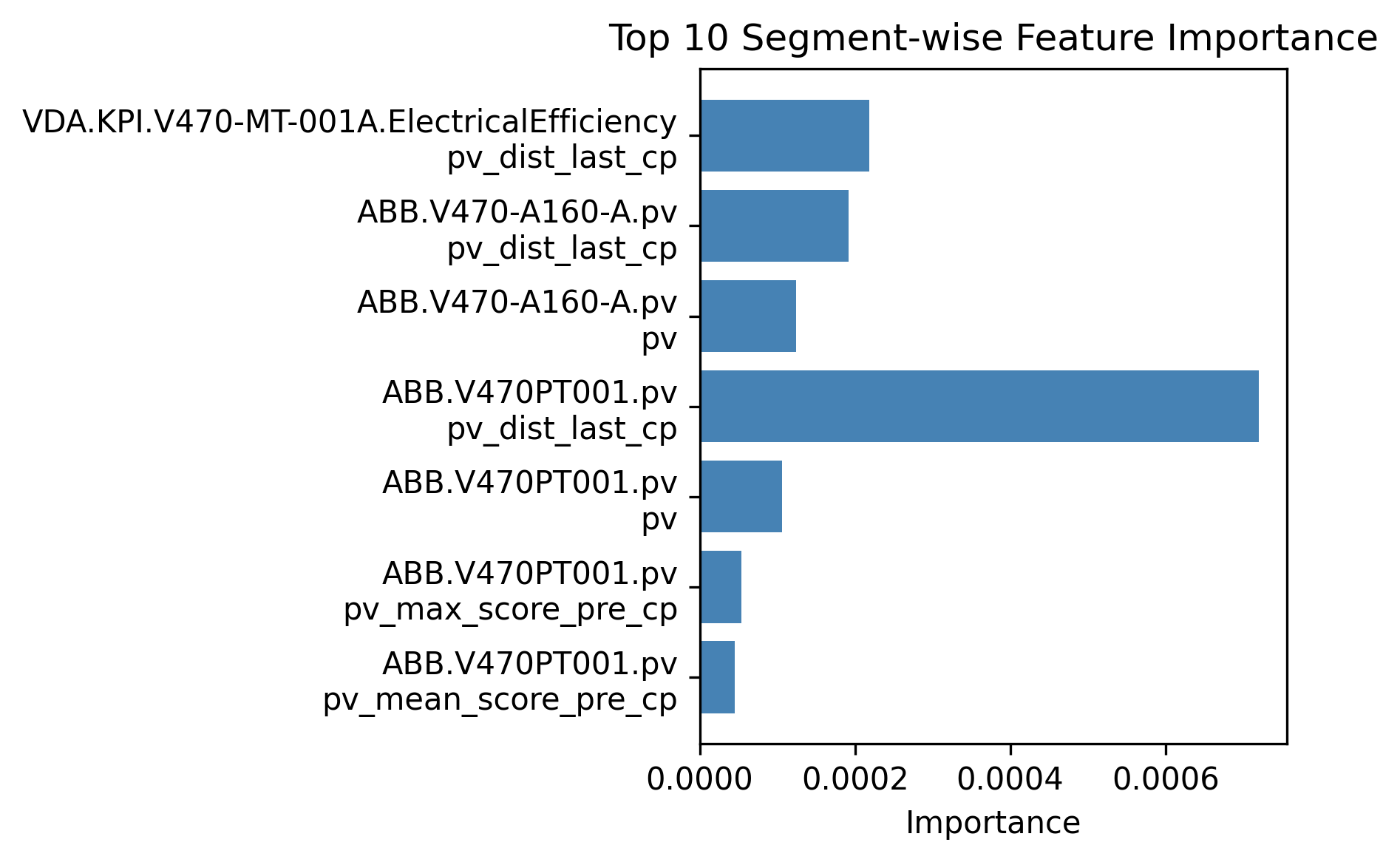}
    \caption{Top 10 segment-wise feature importance values from permutation analysis. The feature pv\_dist\_last\_cp in segment COVA.ABB.V470PT001.pv emerges as the most informative, confirming the relevance of proximity-based and pre-change point metrics in model discrimination.}
    \label{fig:placeholder}
\end{figure}

\begin{table}[htbp]
\centering
\caption{Top 10 Global Features by Mean Random Forest Importance}
\footnotesize
\resizebox{\columnwidth}{!}{
\begin{tabular}{p{6.5cm} r}
\hline
\textbf{Feature} & \textbf{Importance (Mean RF)} \\
\hline
COVA.ABB.V470-A160-A.pv\_segment & 0.1042 \\
COVA.ABB.V470-A041-A.pv\_segment & 0.1030 \\
COVA.ABB.V470-A160-A.pv & 0.0956 \\
COVA.ABB.V470PT001.pv\_segment & 0.0899 \\
COVA.ABB.V470-A165-A.pv\_segment & 0.0895 \\
COVA.ABB.V470-A160-A.pv\_mean\_score\_pre\_cp & 0.0891 \\
Eni.VDA.KPI.V470-MT-001A.ElectricalEfficiency\_segment & 0.0828 \\
COVA.ABB.V470-A041-A.pv & 0.0762 \\
COVA.ABB.V470-A165-A.pv & 0.0728 \\
COVA.ABB.V470PT001.pv & 0.0626 \\
\hline
\end{tabular}
}
\end{table}

From the Random Forest importance, the most influential variables are those related to the segmented process variables (pv\_segment) of key sensors such as COVA.ABB.V470-A160-A, COVA.ABB.V470-A041-A, COVA.ABB.V470-A165-A, and COVA.ABB.V470PT001. Their prominence suggests that the segmentation strategy effectively captured temporal and structural variability, transforming raw process signals into features that better discriminate between normal and anomalous system states. The appearance of the feature COVA.ABB.V470-A160-A.pv\_mean\_score\_pre\_cp among the top ten further indicates that pre-change point behavior contributes valuable contextual information about the system’s approach to instability. The variable Eni.VDA.KPI.V470-MT-001A.ElectricalEfficiency\_segment also holds high importance, implying that aggregated performance indicators complement localized sensor data in explaining variance relevant to fault dynamics.
The Permutation Importance by segment, although based on a finer granularity, reinforces this interpretation. Features such as COVA.ABB.V470-A160-A.pv\_dist\_last\_cp and COVA.ABB.V470PT001.pv\_dist\_last\_cp appear repeatedly across segments, suggesting that the temporal distance from the last change point carries informative weight, possibly acting as a proxy for degradation cycles or process stabilization intervals. Moreover, variables like COVA.ABB.V470-A165-A.pv\_max\_score\_pre\_cp and COVA.ABB.V470-A041-A.pv retain non-negligible influence, aligning with the global feature importance ranking.
In light of both analyses, it would be advisable to retain, for the final modeling phase, the segmented versions of the main process variables, namely COVA.ABB.V470-A160-A.pv\_segment, COVA.ABB.V470-A041-A.pv\_segment, COVA.ABB.V470-A165-A.pv\_segment, and COVA.ABB.V470PT001.pv\_segment. In addition, their corresponding raw process values (.pv) should be included to allow direct-level interpretability. Derived contextual indicators, such as .pv\_mean\_score\_pre\_cp and .pv\_dist\_last\_cp, which capture proximity to critical transitions, are also important. Finally, the efficiency metric Eni.VDA.KPI.V470-MT-001A.ElectricalEfficiency\_segment should be maintained, as it provides an aggregated system-level descriptor.
The heatmap reported in Figure 4 clearly illustrates that segmented variables dominate in both global and segment-level importance, confirming their central role in capturing temporal and structural variability within the process. Derived contextual indicators show moderate global importance but relatively higher localized contribution, suggesting that they provide complementary, segment-specific information. Conversely, raw process variables exhibit moderate importance across both dimensions, reflecting their relevance as baseline descriptors rather than as primary discriminants. Finally, the system efficiency indicator maintains balanced relevance across both scales, linking local sensor behavior to overall process performance.

\begin{figure}
    \centering
    \includegraphics[width=0.75\linewidth]{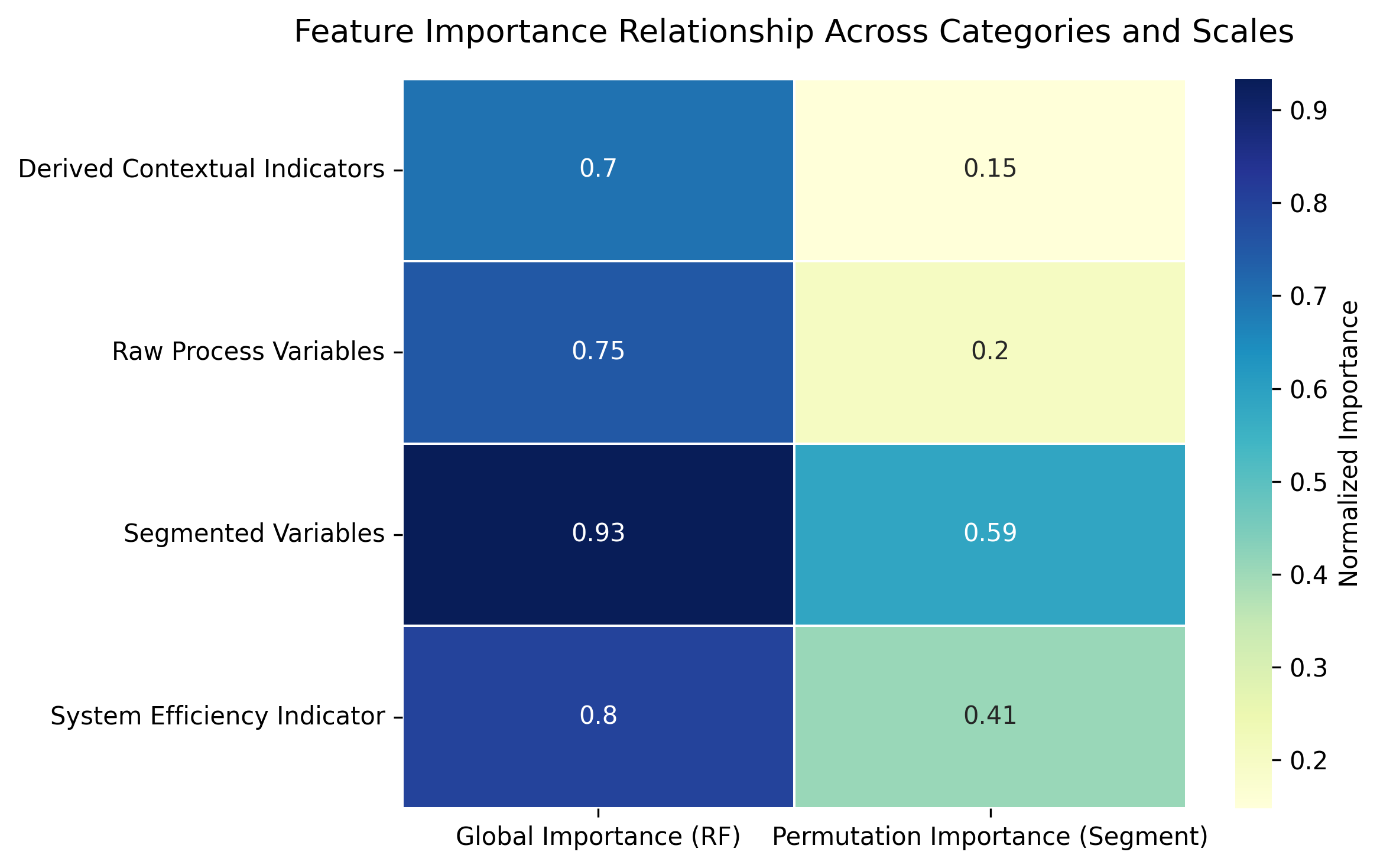}
    \caption{Normalized feature importance by category. 
Comparison of global Random Forest importance and segment-level permutation importance highlights key contributions of segmented variables, raw process variables, derived indicators, and system efficiency.}
    \label{fig:placeholder}
\end{figure}
Together, these features represent a balanced combination of signal segmentation, process-level context, and system-wide performance, offering both interpretability and predictive robustness.
\subsection{Phase 4: Hybrid Model Architectures }
In this phase, the experimental analysis explored several hybrid architectures that combined dimensionality reduction, one-class classification, and tree-based ensemble learning. Specifically, four main configurations were implemented and assessed: (i) PCA + One-Class SVM, (ii) PCA + XGBoost, (iii) One-Class SVM + Random Forest, and (iv) One-Class SVM + XGBoost. These hybrid models were designed to leverage the sensitivity of the One-Class SVM—particularly its ability to capture subtle deviations in high-dimensional data distributions—with the specificity and interpretability of tree-based models, which excel in structured feature spaces and in controlling false positives.
As described in the previous section, prior to hybridization, a feature selection process was conducted to identify the top 10 most informative variables, based on the performance of earlier segmentation-based models. The goal was to determine whether combining complementary learning paradigms could further enhance early anomaly detection, particularly by balancing recall and precision across normal and faulty samples.
Despite the theoretical appeal of these combinations, empirical results revealed that all hybrid configurations performed consistently below the simple ensemble baseline composed of Random Forest and XGBoost trained on the segmented and feature-enriched dataset. For instance, the PCA + One-Class SVM model achieved an AUC-ROC of 0.90 but exhibited poor recall on minority samples and low overall F1-score (0.05), indicating limited discriminative power after projection into the reduced feature space. Similarly, PCA + XGBoost underperformed (AUC-ROC = 0.57, F1 = 0.04), suggesting that the dimensional compression introduced by PCA hindered the downstream learning capacity of the gradient boosting model.
The One-Class SVM + Random Forest and One-Class SVM + XGBoost hybrids also failed to outperform the baseline. While they achieved moderate early detection rates (ETP between 69\% and 91\%) and relatively fast mean Time to Detection (TTD \(\approx 4–12\) samples), their global performance metrics remained inferior to the RF + XGBoost ensemble, which reached an AUC-ROC of 0.98, F1-score of 0.97, and 100\% early detection within the defined window.
Overall, the findings indicate that, in this domain, the marginal gain from combining models with heterogeneous biases does not compensate for the performance loss due to model complexity and sensitivity overlap. The hybrid strategies proved less effective than the ensemble of two tree-based learners operating on a segmented and feature-optimized representation of the data. Consequently, subsequent phases focused on refining segmentation and feature engineering rather than pursuing additional hybridizations.

\section{Experimental Results}
\subsection{The Unbeatable Baseline }
The first set of experiments established a strong reference model against which all subsequent configurations were evaluated. The Random Forest + XGBoost ensemble, trained on the segmented and feature-enriched dataset, consistently delivered the highest overall performance. Specifically, it achieved an AUC-ROC of 0.9760, F1-score of 0.41 for the minority (fault) class, recall of 0.69, precision of 0.29, and an overall accuracy of 0.97.
These results confirm that the combination of segmentation-based data representation and ensemble learning yields a remarkably effective balance between early fault detection and false-alarm control. The ensemble exploits complementary strengths: Random Forest contributes robustness to feature variability, while XGBoost enhances sensitivity to subtle nonlinear patterns. Together, they form a stable and interpretable benchmark for subsequent analyses.
\subsection{The Complexity Penalty}
A comprehensive comparison across all approaches is summarized in Table 5. Despite their conceptual sophistication, more complex or hybrid configurations consistently underperformed relative to the baseline ensemble.

\begin{table}[htbp]
\centering
\caption{Performance comparison of different approaches for anomaly detection (F1 drop in \%)}
\small
\begin{tabular}{p{2.5cm} c c p{2cm}}
\hline
\textbf{Approach} & \textbf{AUC-ROC} & \textbf{F1-Score} & \textbf{F1 Drop (\%)} \\
\hline
Baseline & 0.9760 & 0.41 & Reference \\
CP Features Only & 0.76 & 0.04 & 81 \\
Clustering \(\Delta F\) & 0.54 & 0.04 & 87 \\
PCA + OCSVM & 0.9007 & 0.13 & 68 \\
SVM + RF Hybrid & 0.6475 & 0.06 & 85 \\
Top 10 Features & 0.61 & 0.04 & 90 \\
\hline
\end{tabular}
\end{table}
The results reveal a consistent pattern: increased algorithmic complexity did not translate into improved generalization or discriminative power. In particular, the PCA + One-Class SVM configuration suffered from information loss during dimensionality reduction, leading to weaker separability in the projected space. Similarly, hybrid approaches that combined SVM with Random Forest failed to exploit meaningful complementarity between the two models, suggesting overlapping biases and redundant sensitivity to noise.
Even the model trained on the “top 10” features—selected through Random Forest importance and permutation analysis—exhibited a dramatic decline in predictive capacity. This outcome underscores that isolating features with local discriminative value does not necessarily guarantee global generalization, especially when the data distribution is highly non-stationary or hierarchical in nature.
\subsection{The Trade-off Analysis}
Analysis of metric trade-offs highlights the limits of added complexity. Sophisticated models reached near-perfect recall (\(\sim 99\%\)) but suffered from extremely low precision (2–3\%), generating mostly false alarms. In contrast, the simple ensemble maintained high discriminative ability (AUC-ROC \(\approx 0.98\)) with moderate F1, ensuring reliable early fault detection. Overall, model simplicity combined with informed feature segmentation provides superior stability, interpretability, and practical effectiveness, confirming the ensemble baseline as the optimal approach.

\section{Conclusion}
The experimental results highlight that, in industrial anomaly detection, model simplicity often outperforms complexity. The Random Forest + XGBoost ensemble trained on segmented data consistently achieved superior performance compared to approaches incorporating change point features, clustering-based substructures, or PCA-based hybridizations, which generally did not improve—and sometimes degraded—results. Baseline segmentation alone effectively captured temporal patterns distinguishing normal from anomalous states, while additional engineered features or hybrid learning strategies tended to introduce noise or overlapping sensitivities, reducing generalization in imbalanced, non-stationary datasets. Density-based clustering offered complementary insights but did not enhance supervised model performance, indicating that key substructures were already represented in the segmentation. Overall, the findings suggest that combining simple, interpretable models with domain-informed segmentation provides the best trade-off between accuracy, efficiency, and operational reliability.

\section{ACKNOWLEDGMENT}
This work is supported by ICSC – Centro Nazionale di Ricerca in High Performance Computing, Big Data and Quantum Computing, funded by the European Union NextGenerationEU. The authors gratefully acknowledge Alfonso Amendola and Emilio Villa for their technical and scientific assistance, and Carlo Acutis for his human support, which has been a source of encouragement during this study.


\end{document}